\documentclass[onecolumn]{cohere}
\renewcommand{\fasymbol}{\textcolor{ayadsymbol}{$\vardiamond$}}

\definecolor{green400}{HTML}{0B6623}

\usepackage[utf8]{inputenc}
\usepackage{lipsum}
\usepackage{pdfpages}
\usepackage{colortbl}
\usepackage{tabulary}
\usepackage{longtable}
\usepackage{multirow}
\usepackage{tabularx}
\usepackage{array}
\usepackage{placeins}
\usepackage{tablefootnote}
\usepackage{wrapfig}
\usepackage{graphicx}
\usepackage{amsmath}
\usepackage{cleveref}

\usepackage{pifont}
\definecolor{deeppurple}{HTML}{9e02f7}
\definecolor{forestgreen}{HTML}{2e7d43}

\usepackage{CJKutf8}
\usepackage{arydshln}

\usepackage{xspace} %
\usepackage{makecell} %

\newcommand{\citeposs}[1]{%
  \citeauthor{#1}'s \citeyearpar{#1}%
}

\usepackage[framemethod=TikZ]{mdframed}

\usepackage{multicol}

\usepackage{tcolorbox} %
\tcbuselibrary{most}
\colorlet{LightGreen}{green!20}
\colorlet{LightRed}{red!20}
\colorlet{LightGrey}{black!20}
\tcbset{on line,
        boxsep=1.5pt,left=0pt,right=0pt,top=0pt,bottom=0pt,
        colframe=white,  
        highlight math style={enhanced},
        }

\newtcolorbox{mybox}[2][]{
  colback=white, 
  colframe=lightblue,
  fonttitle=\bfseries,
  coltitle=black,  
  title=#2, 
  #1
}

\definecolor{ayad}{RGB}{148, 156, 229} %
\definecolor{ayadsymbol}{RGB}{76, 110, 230} %
\definecolor{lightblue}{RGB}{211, 227, 252} %
\definecolor{bgblue}{RGB}{247, 250, 255} %
\newcommand*\colourcheck[1]{%
  \expandafter\newcommand\csname #1check\endcsname{\textcolor{#1}{\ding{52}}}%
}
\newcommand*\colourcross[1]{%
  \expandafter\newcommand\csname #1cross\endcsname{\textcolor{#1}{\ding{55}}}%
}

\DeclareSymbolFont{extraup}{U}{zavm}{m}{n}
\DeclareMathSymbol{\vardiamond}{\mathalpha}{extraup}{87}
\definecolor{ayadsymbol}{RGB}{76, 110, 230} %

\colourcheck{blue}
\colourcheck{green}
\colourcheck{red}

\colourcross{blue}
\colourcross{green}
\colourcross{red}
\usepackage{nicematrix}
\usepackage{amssymb}

\setcitestyle{number,square}

\usepackage{epigraph}
\usepackage{comment}
\usepackage{hyperref}

\DeclareUrlCommand\url{\color{black}}

\font\myfont=cmss8 at 20pt

\title{{\myfont The State of Multilingual LLM Safety Research: \\From Measuring the Language Gap to Mitigating It}}

\multiauthors
\author{
    name={Zheng-Xin Yong},
    affiliation={1}
}

\author{
    name={Beyza Ermis},
    affiliation={2},
}
\author{
    name={Marzieh Fadaee},
    affiliation={2},
}
\author{
    name={Stephen H. Bach},
    affiliation={1},
}
\author{
    name={Julia Kreutzer},
    affiliation={2},
}

\affiliations{
    \item[1] Brown University
    \item[2] Cohere Labs
}

\corresponding[*]{Zheng-Xin Yong (\url{contact.yong@brown.edu}), Julia Kreutzer (\url{juliakreutzer@cohere.com})}

\date{\today}
\abstract{ 
This paper presents a comprehensive analysis of the linguistic diversity of LLM safety research, highlighting the English-centric nature of the field. Through a systematic review of nearly 300 publications from 2020--2024 across major NLP conferences and workshops at $^*$ACL, we identify a significant and growing language gap in LLM safety research, with even high-resource non-English languages receiving minimal attention. 
We further observe that non-English languages are rarely studied as a standalone language and that English safety research exhibits poor language documentation practice.
To motivate future research into multilingual safety, we make several recommendations based on our survey, and we then pose three concrete future directions on safety evaluation, training data generation, and crosslingual safety generalization. 
Based on our survey and proposed directions, the field can develop more robust, inclusive AI safety practices for diverse global populations.
\\
\textcolor{BrickRed}{\textbf{Content Warning: This paper contains examples of harmful language.}}
}

\begin{document}
\section{Introduction}
\label{sec:intro}
The rapid advancement of large language models (LLMs) has transformed the artificial intelligence landscape, enabling increasingly sophisticated capabilities across domains including healthcare \citep{singhal2023large,nazi2024large,singhal2025toward}, education \citep{neumann2024edu, zhang2024edu, wen2024edu}, and media content generation \citep{wang2023mediagpt, zhang2024socialmedia,barman2024dark}. 
As these powerful systems are deployed globally and used across different linguistic communities \citep{tamkin2024clio}, ensuring their safe and secure operation across diverse linguistic and cultural contexts has emerged as a critical research imperative. While significant progress has been made in developing safety mechanisms for high-resource languages~\citep{shi2024largelanguagemodelsafety,dong-etal-2024-attacks}, particularly English, the multilingual dimensions of LLM safety remain considerably underexplored.
For example, \textit{all} the public safety evaluation datasets reviewed by \citet{dong-etal-2024-attacks} include English content, with only two datasets being bilingual (English and Chinese).
This gap creates potentially dangerous blind spots in our safety frameworks and raises fundamental questions about the equitable distribution of AI benefits and risks \citep{yong2023lowresource,ermis-etal-2024-one,aakanksha-etal-2024-multilingual,kanepajs2024towards,bengio2025international,peppin2025multilingualdivideimpactglobal}.

\begin{table*}[!t]
\centering
 \resizebox{\textwidth}{!}{
\begin{tabular}{p{.3\linewidth}p{.5\linewidth}p{.2\linewidth}}
\toprule
Categories & Definitions & Examples \\
\midrule
Jailbreaking attacks & Work on designing adversarial prompts to bypass refusal safety guardrails or detecting jailbreaking attacks & \citet{zeng-etal-2024-johnny}, \citet{wang-etal-2024-reinforcement-learning} \\

Toxicity and bias & Work on toxic content and stereotypical bias in training data and output generations & \citet{zhu-etal-2024-plms}, \citet{kim-etal-2024-lifetox} \\

Factuality and hallucination & Work on nonsensical, unfaithful, and factually incorrect content generated by LLMs & \citet{pal-sankarasubbu-2024-gemini} \\

AI privacy & Work on memorization, private data leakage, and unlearning & \citet{dou-etal-2024-reducing}, \citet{shi-etal-2024-ulmr} \\

Policy & Work on governance frameworks, regulatory approaches, and ethical guidelines for responsible AI deployment & \citet{goanta-etal-2023-regulation} \\

LLM alignment & Work that spans multiple subtopics above or is related to other LLM safety subtopics such as RLHF alignment algorithms & \citet{wang-etal-2024-fake}, \citet{yang-etal-2024-self} \\

Not related to safety & Work that does not belong to any of the topics above & \citet{manino-etal-2022-systematicity} \\
\bottomrule
\end{tabular}
}
\caption{Taxonomy for our LLM safety survey study.}
\label{tab:safety_subtopic}
\end{table*}

\begin{CJK*}{UTF8}{bkai}
Multilingual LLM safety encompasses challenges that extend well beyond the simple translation of existing safety techniques. Languages differ not only in their vocabulary and grammatical structures but also in their cultural connotations~\citep{hoijer1954culture,jiang2000culture,everett2012culture,kramsch2014culture,mazari2015culture}, metaphorical expressions~\citep{saygin2001processing,khoshtab-etal-2025-comparative}, taboos~\citep{dewaele2004emotional}, and social norms~\citep{sridhar1996societal,baquedano2007growing,fasya2021sociocultural}. Therefore, content that is harmless in one cultural context may be deeply offensive or harmful in another \citep{keipi2016online,ermis-etal-2024-one,aakanksha-etal-2024-multilingual,korre-etal-2025-untangling}, or vice versa. For instance, in South-East Asia, the term ``\textit{banana}''---which connotes ``yellow on the outside, white on the inside''---is used to disparage people of Asian descent who are perceived as forgoing their cultural identity and having adopted Western cultural values and behaviors~\citep{khoo2003banana,trieu2019understanding}. On the other hand, the Chinese word 屌, which literally translates as ``dick'', can be used in both offensive (i.e., swear words) and non-offensive (i.e., an adjective to praise someone who possesses a remarkable talent) settings \citep{carson2021offensive}.
\end{CJK*}

The wide disparity in language resources \citep{joshi-etal-2020-state,nigatu-etal-2024-zenos}––from high-resource languages like English, Mandarin, and Spanish to thousands of low-resource languages––creates uneven safety landscapes with potentially severe consequences for marginalized linguistic communities. Several commercial LLMs have demonstrated significantly weaker safety performance when prompted in non-English languages, producing harmful content and undesirable outputs that would be filtered in English contexts \citep{yong2023lowresource,deng2024multilingual,wang-etal-2024-languages,ghanim2024arabicjailbreak,yoo2024code,he2024tuba,shen2024language,nigatu2024searched,poppi2025towards,aakanksha-etal-2024-multilingual,jain2024polyglotoxicityprompts,chan2025speak}. 
These disparities in safety protections, combined with increasingly capable LLMs, risk magnifying societal harms within multilingual communities.
While companies behind frontier LLMs have taken concerted efforts to perform multilingual safety alignment training and red-teaming \citep{grattafiori2024llama,cohere2025command,openai2025gpt45}, these initiatives remain limited in scope.
For instance, among the top-ranking LLMs on Chatbot Arena––a widely used leaderboard platform for evaluating LLMs through user-submitted preference––20 of 24 of those that provide a system report have wide multilingual support, but \textit{only 5} reported multilingual safety alignment training and red-teaming efforts.
This \textit{gap between multilingual deployment capabilities and safety alignment} calls for further participation from both private enterprises and academia on multilingual safety alignment.

We perform a systematic review of nearly 300 LLM safety publications over the past five years in ACL proceedings (\Cref{sec:language-gap}), and we uncover a concerning trend: the vast majority of safety research is centered on English-language models, while comparatively little work addresses safety in non-English or multilingual contexts. This imbalance has become more pronounced over time.
Even Mandarin Chinese––the second most studied language––still has about \textit{ten times less} research than English. This disparity persists across multiple subdomains of safety research. Furthermore, non-English languages are rarely studied as a standalone language but rather as part of broader multilingual evaluations, which often lack the nuance and depth necessary to address language-specific safety challenges and cultural contexts. Lastly, we discover that only half of English safety research publications document the limitations of their language coverage.

These findings highlight critical gaps in the current landscape of LLM safety research and motivate the need for more targeted efforts to address multilingual safety concerns.
To help close this gap, we outline three tractable directions for future multilingual safety work: (1) developing culturally grounded evaluation benchmarks, (2) curating diverse multilingual safety training data, and (3) deepening our understanding of alignment challenges across languages.

\begin{figure*}[!t]
    \centering
    \includegraphics[width=0.8\linewidth]{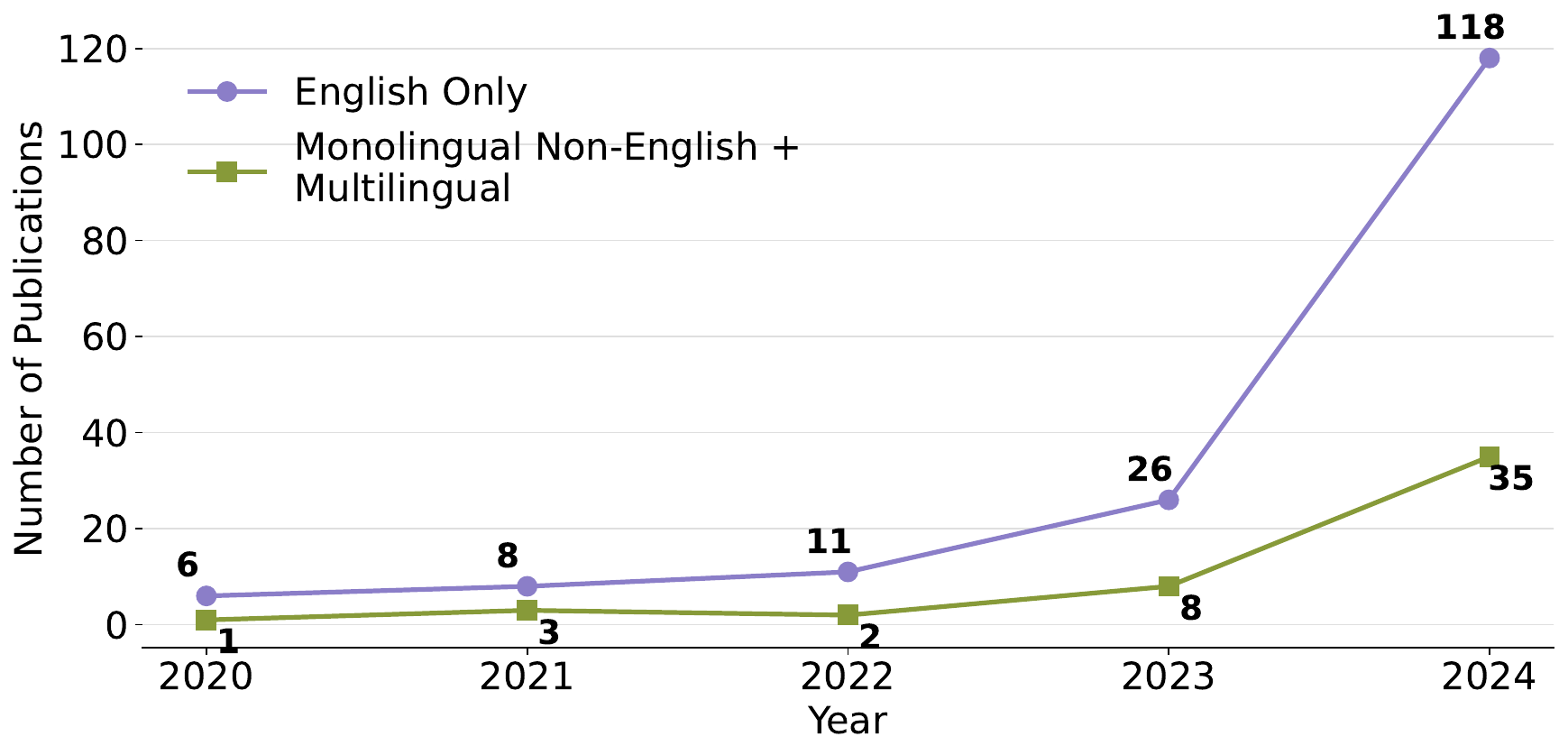}
    \caption{Trends of English-only and multilingual LLM safety publications in $^*$ACL conferences and workshops over the past five years: the language gap in LLM safety research widens. 
    }
    \label{fig:pub-by-year}
\end{figure*}

\section{The Language Gap in LLM Safety Research}\label{sec:language-gap}
To understand the language gap in LLM safety research, we systematically survey relevant papers and analyze how safety research is distributed across languages and subtopics, as well as how non-English language research is conducted and reported.

\subsection{Methodology}
We collect work related to LLM safety and manually annotate the languages studied in each paper, along with their safety subtopic. To reduce human annotation efforts while ensuring that our findings reflect the overall trends in the field, we perform the following strategies: 
\begin{enumerate}
    \item \textbf{Venue selection}: We focus on all $^*$ACL venues such as ACL and EMNLP, including both conferences and workshops, as we believe they are the venues with the most linguistically diverse NLP works compared to other venues such as ICLR, NeurIPS, and ICML.
    \item \textbf{Keyword filter}: We filter the safety-related publications through keyword matching with words ``safe'' and ``safety'' in paper abstracts. Using these two terms
    we get a good proxy for the distribution of diverse LLM safety literature.  
    \item \textbf{Manual categorization}: We adopt a simplified taxonomy following \citet{cui2024risk}, which is representative of the type of safety work published at $^*$ACL, and we manually categorize publications into seven different subtopics as shown in \Cref{tab:safety_subtopic}.

    \item \textbf{Language Documentation}: We annotate the languages that each work addresses,\footnote{If the languages studied were not explicitly mentioned, we followed up on their training and evaluation datasets to identify the language coverage of the work.} and we indicate if the language(s) studied are mentioned in the work. We group them into three categories: monolingual English, monolingual non-English, and multilingual (covering two or more languages).
\end{enumerate}

\begin{wraptable}{r}{.5\columnwidth}

\centering
\begin{tabular}{llcc}
\toprule
\textbf{Annotation Task} & \textbf{Type} & \textbf{Avg} & \textbf{Std} \\
\midrule
  Safety topic & Categorical & 0.83 & 0.19\\
  Has non-English? & Binary  & 0.81 & 0.15\\
  Specifies languages? & Binary & 0.80 & 0.04 \\
  Covered languages & List & 0.96 & 0.05\\
\bottomrule
\end{tabular}
\caption{Average and standard deviation of agreement between four pairs of annotators. 
Agreement on `language coverage' is measured with Jaccard similarity, and all other categories are measured with Cohen's $\kappa$.}
\label{tab:agreement}

\end{wraptable}

Annotations were manually performed by the authors. In total, we annotated nearly 300 publications from year 2020 till year 2024. Of these, 28\% were false positives from our keyword matching process (i.e., unrelated to LLM safety), and were filtered out before we perform further analysis.\footnote{We release our annotations on \url{https://huggingface.co/CohereLabsCommunity/multilingual_safety_survey2025}.}

\Cref{tab:agreement} reports the mean and standard deviation of pairwise inter-annotator agreement scores on subsets of 20 repeated annotations. We perform a $4\times20$ pairwise agreement study across distinct subsets to maximize the representativeness of our survey corpus and ensure robust assessment of annotation consistency. 
We find that inter-annotator agreement is consistently high, between 0.80 and 0.96 on average per category, but we note that the annotations may still contain imperfections.

\begin{figure*}[!t]
    \centering
    \includegraphics[width=0.95\linewidth]{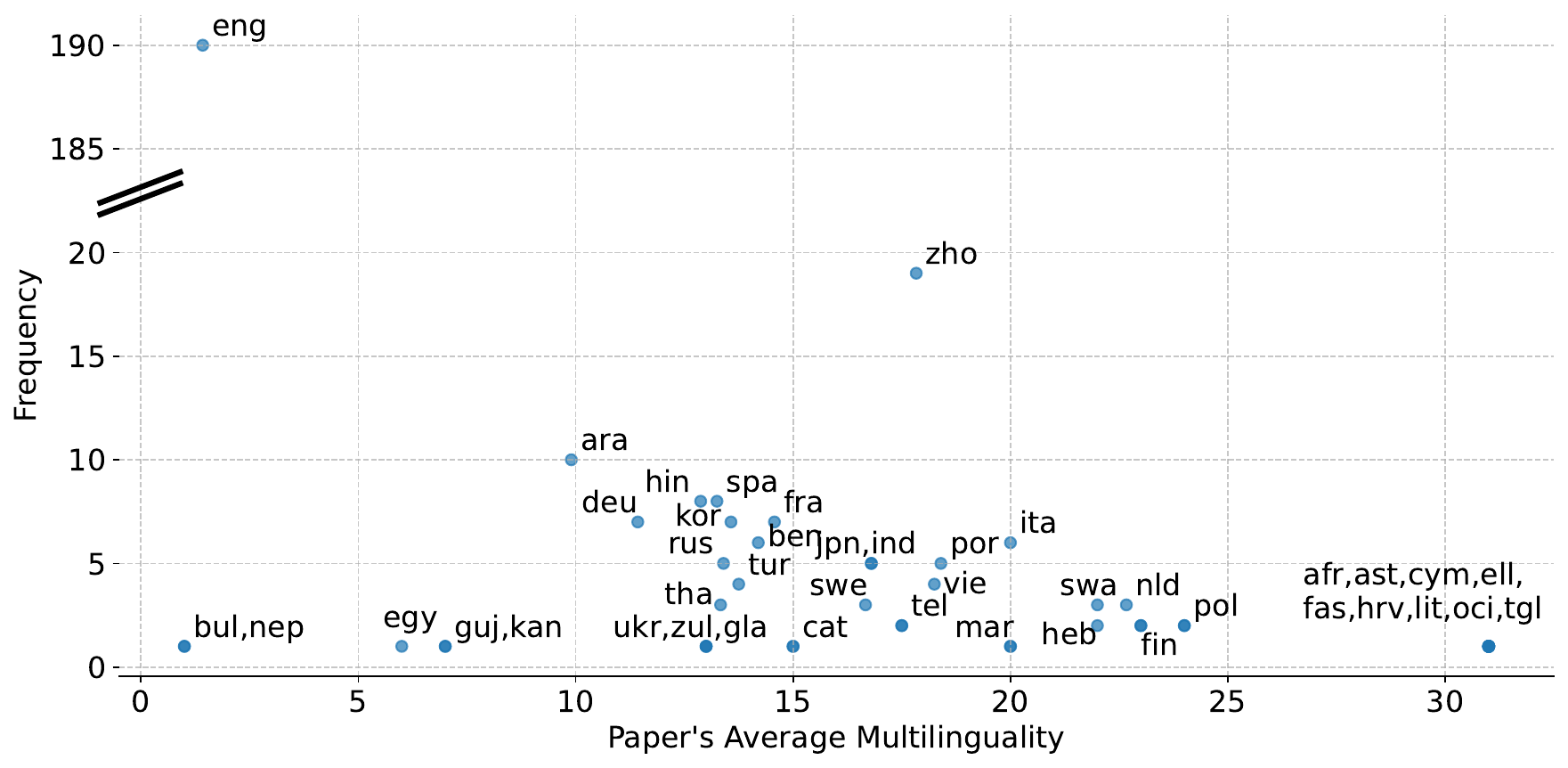}
    \caption{Measure of how often a language is studied (``Frequency'') and the average number of languages covered by all papers in which the language appear in (``Paper's Average Multilinguality'').
    }
    \label{fig:lang_dist}
\end{figure*}

\subsection{Findings} 
\textbf{English-centricity of LLM safety research.}
\Cref{fig:pub-by-year} highlights a stark language imbalance in LLM safety research published at $^*$ACL conferences and workshops over the past five years. The data reveals a clear English-centric pattern that has persisted throughout this period. English-only research dominates across all years, with a particularly dramatic increase in recent publications. The trend shows consistent underrepresentation of multilingual non-English research, with the gap widening significantly over time. While both categories have grown as LLM safety has gained prominence, the proportional imbalance remains. 
English-only publications have consistently outnumbered multilingual and non-English work, and this absolute gap has widened over time, from 5 in 2020 to 83 in 2024. While both categories have grown, the increase is disproportionately concentrated in English-only research.

\textbf{Non-English languages are studied in herds.} Another aspect of the marginalization of non-English languages is that they are often addressed as part of large multilingual evaluations, rather than studied in depth on their own. In many cases, breadth is prioritized over depth, and multilingual studies are preferred over focused analyses of monolingual ones.\footnote{Since our study only captures published papers, we might be missing out on rejected works. There may be a reviewer preference for multilingual over monolingual non-English papers.}
This is shown in \Cref{fig:lang_dist} which provides a detailed breakdown of how frequently a language is studied (y-axis) and how often it is studied alongside other languages (x-axis). English (\texttt{eng}) exhibits overwhelming dominance with a frequency nearly ten times higher than Chinese (\texttt{zho})––the second most studied language.
However, English is primarily studied in isolation, resulting in a low average multilinguality score.

In contrast, languages with moderate representation like Chinese (\texttt{zho}), Arabic (\texttt{ara}) and Spanish (\texttt{spa}) appear primarily in multilingual studies, suggesting that deeper, language-specific safety analyses remain limited even for widely spoken languages. This trend is even more noticeable for under-resourced languages such as Swahili (\texttt{swa}) and Telugu (\texttt{tel}), and especially for languages at the extreme end of the multilingualism spectrum such as Afrikaans (\texttt{afr}), which appears only in a single paper that covers approximately 30 languages \citep{guerreiro2023hallucinations}. Such inclusion severely limits the possibility for language-specific safety analysis and gaining meaningful insights. 
We commend focused analysis on individual lower-resource languages such as \citet{nakov2021covidbulgarian} and \citet{niraula2021offensivenepali}, who specifically study disinformation and offensive language detection in Bulgarian and Nepali social media, respectively.

\begin{figure*}[t!]
    \centering
    \begin{subfigure}[t]{0.4\textwidth}
        \centering
        \includegraphics[width=\textwidth]%
        {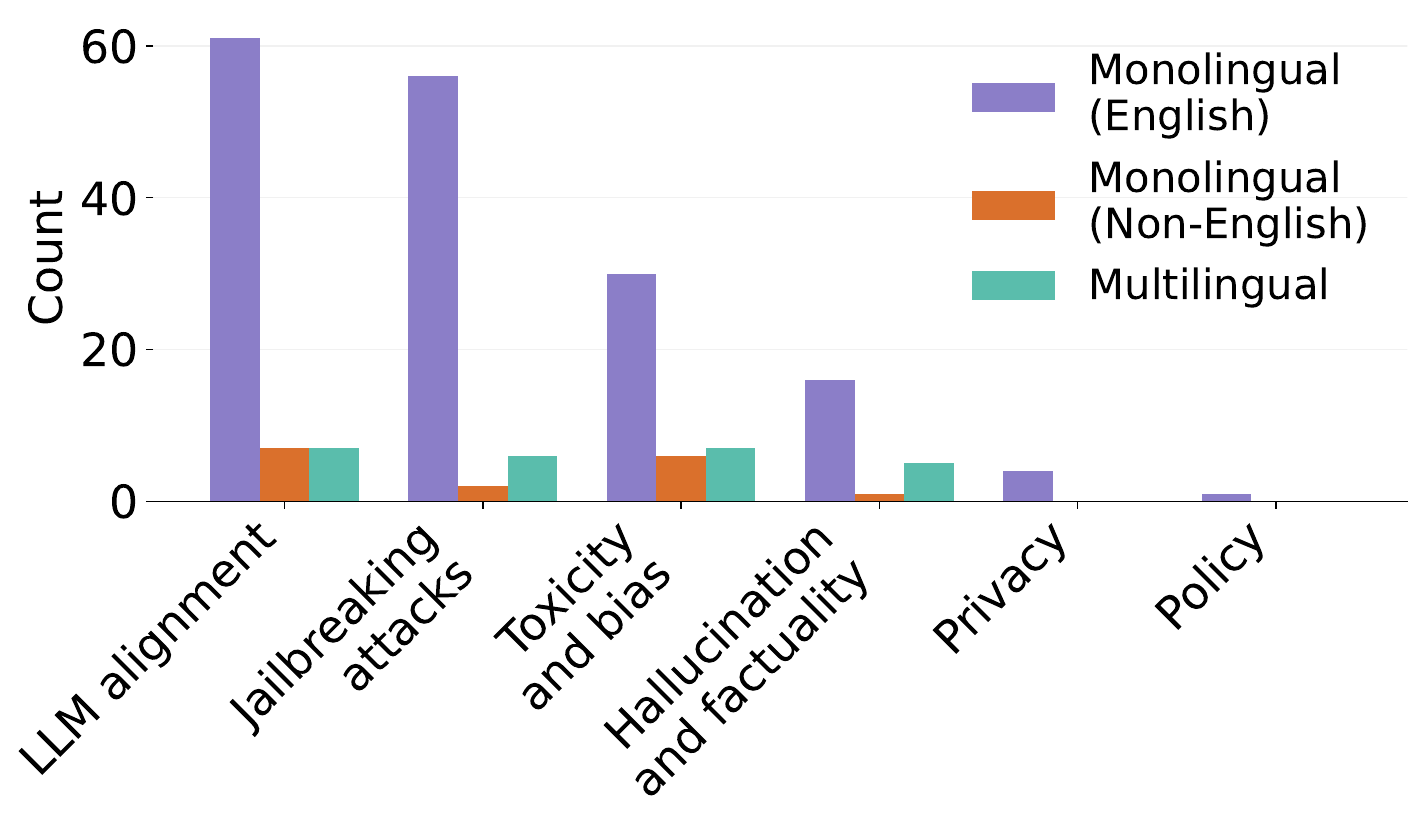}
        \caption{Topic distribution}
    \end{subfigure}%
    \hfill
    \begin{subfigure}[t]{0.5\textwidth}
        \centering
        \includegraphics[width=\textwidth]%
        {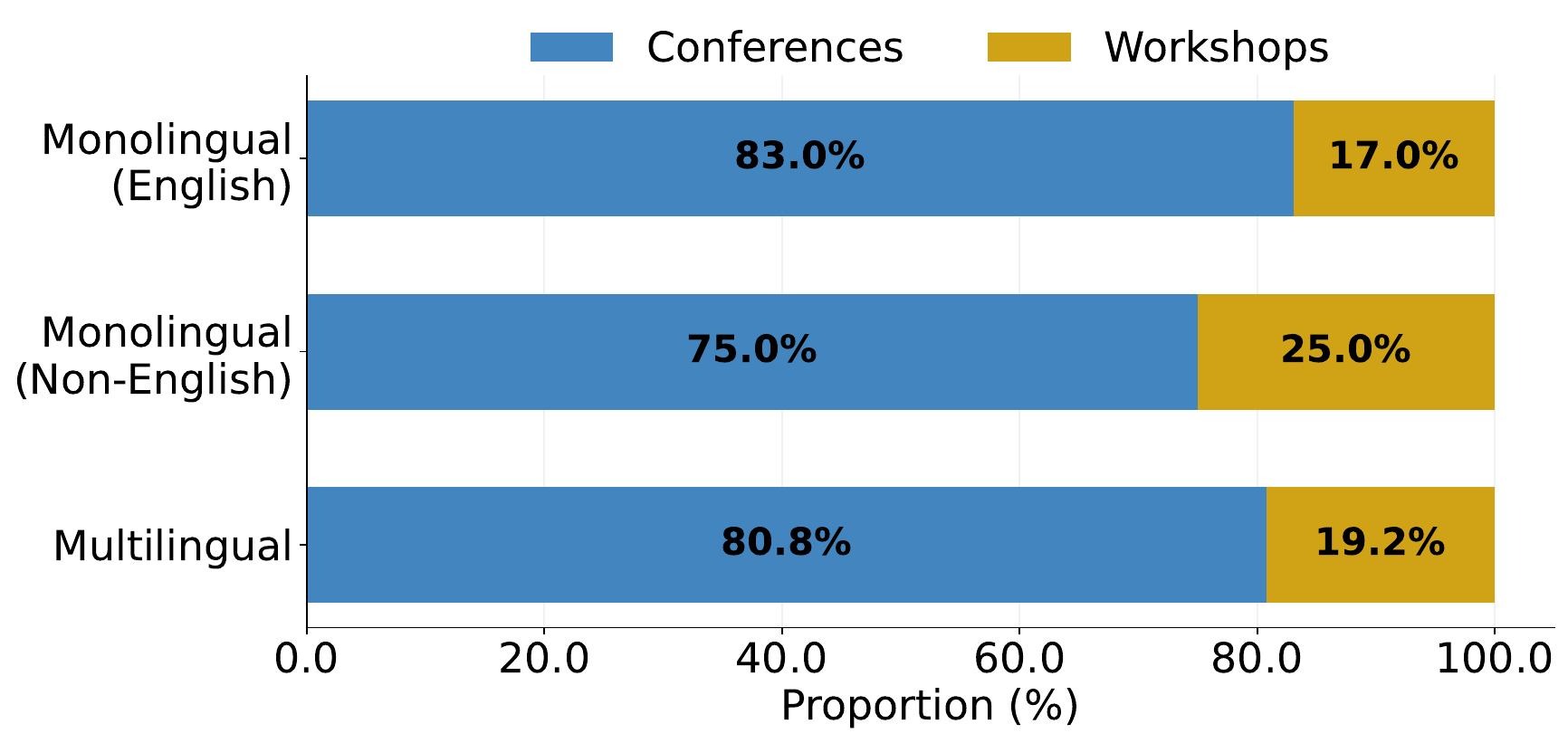}
        \caption{Venue distribution}
    \end{subfigure}
    \caption{Distribution of LLM safety publications by (a) safety subtopics and (b) publication venues.}
     \label{fig:topic-venue-breakdowns}
\end{figure*}

\textbf{Disparities in subtopics of safety.} 
Breaking down LLM safety publications by specific safety subtopics in \Cref{fig:topic-venue-breakdowns}(a), we find that English-centricity persists across all domains, with English-only publications substantially outnumbering multilingual work in every category.
LLM alignment and jailbreaking attacks demonstrate the most pronounced disparities, suggesting that these critical safety areas receive particularly limited cross-linguistic attention. In particular, LLM alignment work involving evaluation \citep{yuan2024rjudge,hua2024trustagent,hammoud2024modelmerging,gabriel2024aimentalhealth} and algorithmic improvement \citep{zhou2024beyond,hassan2024activelearning} would benefit from further research with expanded language coverage. Toxicity and bias research shows a similar pattern despite being a domain where cultural and linguistic variations are especially relevant \citep{costajussa2023multilingual,tao2024cultural,devinney2024dont,bhutani2024seegull}. The near absence of multilingual work in privacy and policy domains indicates these emerging safety concerns are being conceptualized almost exclusively through an English-language framework, potentially overlooking important cultural and legal variations that exist across different linguistic contexts \citep{larsen2024aivaluealignment}.

\textbf{Valuable role of workshops.} \Cref{fig:topic-venue-breakdowns}~(b) reveals an interesting pattern in the distribution of LLM safety publications across venue types. 
While conferences dominate across all language categories, monolingual non‑English safety papers are 46\% relatively more likely to appear in workshops than English‑only papers, highlighting the valuable accessibility that workshops offer for this line of work.
This suggests that non-English safety research faces a higher barrier to entry at prestigious conferences, whereas workshops, such as Workshop on Gender Bias in Natural Language Processing (GeBNLP) and Workshop on Safety for Conversational AI (Safety4ConvAI), serve as more accessible venues for disseminating non-English safety research.
The pattern indicates that, beyond the overall English-centricity of safety research documented in previous figures, additional structural factors may be affecting how non-English safety work is evaluated and disseminated within the community.

\textbf{Language documentation practice differs for English-only research.}
We argue that it is important for LLM safety research to \textit{explicitly document the languages studied} (also known as Bender's rule \citep{bender2011achieving,bender2019rule}) for two key reasons. (1) Safety alignment does not necessarily generalize across languages \citep{yong2023lowresource,wang2024alllanguages,yoo2024code,ghanim2024arabicjailbreak}. 
Clearly stating which languages were included enables future researchers to understand the specific linguistic contexts in which safety findings have been validated.
(2) By explicitly acknowledging language limitations, the field can more accurately measure progress in expanding safety coverage across languages, thus encouraging a more equitable distribution of safety research to serve a broader range of global populations. 

\begin{wraptable}{r}{.5\columnwidth}
\centering
\resizebox{.5\columnwidth}{!}{
\begin{tabular}{lcc}
\toprule

\textbf{Category} & \multicolumn{2}{c}{\textbf{Does the paper mention languages studied?}} \\
& No (\textcolor{red}{$\downarrow$}) & Yes (\textcolor{green400}{$\uparrow$}) \\
\midrule
Mono. English & 50.6\% & 49.4\% \\
Mono. Non-English & 0.0\% & 100.0\% \\
Multilingual & 0.0\% & 100.0\% \\
\bottomrule
\end{tabular}
}
\caption{Proportion of language documentation practice among LLM safety publications.}
\label{tab:bender_rule}
\end{wraptable}

Based on the data presented in \Cref{tab:bender_rule}, we observe substantially different patterns in language documentation practices across LLM safety publications. English-only publications show a concerning trend with 50.6\% failing to explicitly name the language studied --  in other words, ``English'' is not mentioned throughout the paper. In contrast, both non-English monolingual and multilingual publications demonstrate full compliance, with 100\% explicitly documenting the languages studied. This disparity highlights a systematic bias in reporting practices, where English-centered research often proceeds under an implicit assumption of universality, whereas non-English research demonstrates greater methodological transparency.

\subsection{Moving Forward for $^*$ACL Venues}
Our survey reveals that English safety research remains overwhelmingly dominant in nearly every dimension—publication volume, topical coverage, methodological reporting, and conference visibility. Nonetheless, \Cref{fig:pub-by-year} shows an encouraging trend of growing multilingual safety research over time. 
One concrete and low-effort step toward improving documentation is integrating language coverage reporting into $^*$ACL proceedings. OpenReview submissions already include a metadata field where authors can indicate the languages studied, but this information is currently private. Making this metadata public would allow for more transparent tracking of linguistic representation and support future meta-analyses of multilingual research, particularly in the context of LLM safety.

Addressing the deeper structural imbalance in language and topic representation will require long-term efforts.
We believe that conference and workshop organizers can provide incentive structures to address this systemic imbalance, such as special conference theme tracks dedicated to multilingual safety subtopics and/or creating shared workshop tasks on multilingual safety benchmarks.
These initiatives could meaningfully expand the scope and visibility of research beyond English, helping the community better serve diverse user populations.

\section{Future Research Directions for Multilingual LLM Safety}

In addition to providing recommendations to $^*$ACL organizers, we propose several key research priorities for researchers and model developers to advance multilingual LLM safety alignment.

\subsection{Safety Evaluation for Multilingual Models} 

\begin{table*}[htbp]
  \centering
  \small
  \resizebox{\textwidth}{!}{%
  \begin{tabular}{lcccccccccccc}
    \toprule
    \textbf{Models} & \textbf{en} & \textbf{zh} & \textbf{fr} & \textbf{ru} & \textbf{de} & \textbf{ar} & \textbf{hi} & \textbf{es} & \textbf{ja} & \textbf{bn} & \textbf{Average} $\uparrow$ & \textbf{Worst Case}$^*$ $\uparrow$\\
    \midrule
    ChatGPT \citep{openai2022chatgpt} & 99.0 & 91.9 & 86.3 & 87.5 & 85.3 & 90.8 & 81.7 & 91.5 & 79.0 & 62.6 & \textbf{85.56} & 62.6 \\
    PaLM-2 \citep{anil2023palm} & 89.7 & 78.4 & 84.6 & 85.9 & 83.6 & 82.6 & 83.0 & 85.7 & 70.1 & 78.1 & 82.17 & \textbf{70.1} \\
    Llama-2 \citep{touvron2023llama} & 85.4 & 73.5 & 83.2 & 82.3 & 82.0 & - & 63.5 & 79.3 & 71.0 & - & 77.53 & 63.5 \\
    Vicuna \citep{vicuna2023} & 94.0 & 89.4 & 90.6 & 83.3 & 88.3 & 43.4 & 36.8 & 88.8 & 60.2 & 18.4 & 69.32 & \textcolor{BrickRed}{\textbf{18.4 (!)}} \\
    \bottomrule
  \end{tabular}
  }
  \caption{Harmlessness scores of different models across 10 languages, based on the results from \citep{wang-etal-2024-languages}. We augment the original table with a new \textit{"Worst Case"} column for the lowest harmlessness score. We use \textbf{bold text} to indicate the cases where average score is not necessarily aligned with worst-case score, and we use \textcolor{BrickRed}{\textbf{red text and exclamation mark}} to indicate how not reporting Worst-Case score can create a false sense of safety.}
  \label{tab:avg_worst-case_safe}
\end{table*}

\textbf{Moving beyond average safety criterion.}
Traditional evaluation metrics focus on average performance across languages, for which the model that maximizes the uniformly weighted average across tasks and languages is considered best.
However, this criterion is susceptible to outliers (e.g., due to unsupported languages) and not suitable for comparing models with different language and task support~\citep{kreutzer2025d}. 
In the context of multilingual safety, where reporting average scores is the norm \citep{guan2024deliberative}, this matters even more since averaging might obscure critical safety failures. 

To illustrate this blind spot, we add the additional \textit{worst-case} harmlessness score metric to an ACL 2024 paper \cite{wang-etal-2024-languages} and report the results in \Cref{tab:avg_worst-case_safe}. The table reveals two findings. First, if the winner were chosen based solely on the highest average harmlessness score, it would be ChatGPT \citep{openai2022chatgpt}, with a score of 85.56. However, its worst-case score (i.e., the lowest harmlessness score across languages) is only 62.6. This is notably lower than the worst-case score of PaLM-2 (70.1), despite PaLM-2 \citep{anil2023palm} having a lower average score (82.17). This discrepancy highlights that strong average performance does not necessarily reflect robustness in the worst-case scenarios. Second, and more importantly, despite a high average harmlessness score, Vicuna's \citep{vicuna2023} worst-case harmlessness score is just 18.4 due to unsafe behaviour in Bengali (bn). 
This suggests that relying on average metrics alone may create a false sense of safety, potentially leading to the deployment of models like Vicuna in languages where they produce harmful content. In future work, we believe that, in addition to reporting worst-case performance to ensure that models meet fundamental safety thresholds across all languages, researchers should explore designing adaptive thresholding mechanisms that establish language-specific safety baselines according to their unique cultural contexts and user groups.

\textbf{Wider language coverage in evaluation.} We observe that current multilingual red-teaming practice mostly focuses on languages that models are finetuned on during post-pretraining processes, such as instruction-following and alignment finetuning \citep{ustun-etal-2024-aya,grattafiori2024llama}.
Given that language contamination in pretraining can facilitate crosslingual transfer \citep{blevins-zettlemoyer-2022-language}, it raises valid concerns about whether \textit{exempting} certain languages from the safety evaluation of multilingual LLMs is justified. Language exemptions risk creating blind spots in safety assessments precisely where they might be most needed, as models can bypass safety guardrails when prompted in languages underrepresented in pretraining \citep{shen2024language}. For instance, the Llama-3 model report presents red-teaming results for only eight languages (six of which are high-resource) \citep{grattafiori2024llama}. Yet the strong multilingual model has been adapted for languages not covered in its safety evaluation, such as Indonesian \citep{huang2024meralion}.

We urge researchers to develop more sophisticated evaluation protocols that can detect and account for potential contamination and to issue disclaimers when safety alignment has not been conducted in certain languages. This would help ensure that speakers of those languages are aware of potential risks. Such transparency would allow communities to make informed decisions about model deployment while encouraging greater accountability from developers to expand alignment efforts to underserved languages.

\textbf{Incorporate diverse and natural linguistic patterns.} We believe evaluating multilingual safety requires a fundamental shift away from treating evaluation as merely adding more languages to existing benchmarks, as they should incorporate linguistic patterns used by real-life speakers. One case study is \textit{code-switching}––the communication pattern of alternating between languages within a single utterance \citep{nilep2006code,gardner2009code,winata-etal-2023-decades}––which is shown to be able to jailbreak multilingual safety guardrails \citep{yoo2024code,yang2024benchmarking,song-etal-2025-multilingual}. Another example is \citeposs{ghanim2024arabicjailbreak} discovery that while LLMs remain safe in standardized Arabic scripts, they are jailbroken when Arabic inputs are written in Arabizi form––a system of writing Arabic using English characters and commonly used among native speakers communicating digitally \citep{yaghan2008arabizi}. These examples show that current safety evaluation frameworks that predominantly evaluate languages in a monolingual setting fail to capture the complex reality of multilingual communication. Future work on multilingual red-teaming should develop a methodology that systematically accounts for diverse multilingual multi-turn interactions among users \citep{li2025beyond} to ensure that models remain safe across the full spectrum of real-world usage patterns rather than just in artificial monolingual test scenarios.

\subsection{Culturally-Contextualized Synthetic Training Data}
Collecting labeled training data for LLM safety alignment can be resource-intensive, and many English-centric research has turned to using synthetic data generation \citep{bai2022constitutional,kruschwitz-schmidhuber-2024-llm,samvelyan2024rainbow}. However, exploration of multilingual synthetic safety data has been relatively underexplored. Here, we propose two viable future research directions based on \textit{constitutional AI} framework \citep{bai2022constitutional,kundu2023specific} for cultural contextualization~\citep{qiu2025multimodal, guo2025care, qiu2024evaluating, yin2024safeworld}

\textbf{LLM Generation.} Under constitutional AI framework, LLMs are first prompted to generate harmful (or harmless texts). They are then presented with a set of human-written principles that capture culture-specific harms so that they can engage in a multi-turn process of critiquing and revising originally harmful generations to harmless generations (or vice versa), to create culture-specific preference pairs for alignment training. 
Enabling constitutional AI for multilingual and multicultural alignment data generation requires close collaboration among linguists, cultural anthropologists and AI researchers to co-create three key components: 
(1) culturally-informed constitutional principles that reflect diverse value systems and ethical frameworks across different societies \citep{kirk2024prism,pistilli2025civics}; 
(2) sufficiently capable multilingual LLMs that can both understand these principles and generate high-quality content in target languages \citep{qin2024multilingual,huang2024survey}; and 
(3) evaluation protocols involving native speakers and cultural experts to validate both the constitutional principles and the resulting synthetic data \citep{Kyrychenko2025c3ai}.
This direction offers a pathway toward scalable, culturally grounded alignment practices that make LLM safety more inclusive and globally relevant.

\textbf{Machine Translation.} 
Machine translation (MT) often fails to capture or preserve culture-specific harms and may introduce undesirable societal biases such as gender stereotyping \citep{savoldi2021gender,ahn-etal-2022-knowledge,wang-etal-2022-measuring,costa-jussa-etal-2023-multilingual,costa-jussa-etal-2023-toxicity}. The iterative refinement process from the constitutional AI framework can detect and mitigate translation artifacts that might inadvertently encode harmful content or lose important cultural nuances. Unlike direct LLM generation, this approach can take advantage of the decades-long research in MT, especially on cross-cultural adaptation studies \citep{maxwell1996translation,de2018systematic,gorecki2014language,mbada2015translation, pilz2014brazilian}.
Future work should focus on developing automated methods to identify culture-specific safety issues that might be lost in translation, especially for languages with limited digital presence and linguistic resources.

\subsection{Towards Understanding Crosslingual Safety Generalization}

Most existing safety alignment data are centered on English or Chinese \citep{rottger2025safetyprompts,costa-jussa-etal-2024-overview,derczynski-etal-2024-countering}. It is important to understand how safety alignment generalizes across languages, so the model developers can \textit{anticipate} potential failure modes when alignment training data lack language coverage.

\textbf{Mechanistic interpretability.} This scientific approach of reverse-engineering neural networks to understand precisely how they process information at the circuit and component levels–––allows researchers to characterize mechanisms that enable or prevent safety alignment knowledge transfer. We believe this research direction is particularly helpful in explaining several phenomena, such as why detoxification and debiasing can transfer effectively across languages \citep{li-etal-2024-preference,reusens-etal-2023-investigating} but not refusal training \cite{shen2024language,aakanksha-etal-2024-multilingual,wang2025refusaldirectionuniversalsafetyaligned}, or to what extent safety alignment is \textit{preserved} after language adaptation to underrepresented languages \citep{yong-etal-2023-bloom,lin2024mala,ji2024emma}.
Insights from this research direction can inspire novel training techniques that facilitate zero-shot crosslingual generalization of alignment training and maintain safety consistency as language coverage expands. 

\textbf{Training Data Influence Analysis.} We also recommend exploring the use of influence functions \citep{grosse2023studying,ruis2025procedural} to study crosslingual alignment. 
This technique enables researchers to trace how specific training examples causally affect model behavior during generation.
Training data influence analysis offers a valuable complement to mechanistic approaches for investigating two key open questions. For crosslingual generalization, it can help quantify how safety-relevant examples---especially those from high-resource versus low-resource languages---contribute to harmful or aligned outputs. For language adaptation, influence functions can identify problematic documents within the continued pretraining corpus, enabling more targeted curation of safer language-specific data.
To our knowledge, there is currently \textit{very limited work} on analyzing training-example-to-output relationships for multilingual safety-relevant behaviors. This presents a promising and underexplored direction for improving alignment practices across languages.

\section{Related Work and Discussion}
Our work contrasts prior survey literature on multilingual NLP \citep{joshi-etal-2020-state,pamungkas2023towards,yadav2022survey, winata-etal-2023-decades, huang2024survey,  qin2024multilingual, wu2025bitter} by focusing on LLM safety. The limitations we identify align with concerns by \citet{blasi-etal-2022-systematic} regarding systematic inequalities in language technology, which privileges certain sociolinguistic groups through choices in data collection, annotation protocols, and evaluation. Our findings suggest these inequalities may be even more pronounced in safety research, where cultural and linguistic nuances significantly impact harm and mitigation strategies.

Recent efforts to catalog LLM safety research challenges \citep{barez2025open,debar2024emerging,anwar2024foundational} have primarily centered on threats identified through English-language models, often overlooking multilingual aspects. This gap, along with our survey findings, echoes the ``square-one bias'' phenomenon~\citep{ruder-etal-2022-square}: When NLP researchers moves beyond optimizing for usefulness (e.g., accuracy), their study is often only conducted in a single direction of either safety, interpretability, or multilinguality. This siloed approach means that progress in one dimension rarely informs the others, resulting in a fragmented research landscape where multilingual LLM safety research  remains underdeveloped.

\section{Conclusion}
Our analysis of nearly 300 publications (2020-2024) reveals a significant language gap in LLM safety research, with even high-resource non-English languages receiving minimal attention and typically appearing only in multi-language studies that lack the depth of English-focused work. 
This linguistic imbalance potentially leaves language-specific risks undetected as LLMs deploy globally. 
To address these disparities, we make recommendations to future conferences and highlight several critical future research directions.

\section*{Limitations}

\textbf{Coverage of venues} Due to the focus on $^*$ACL venues, we might have missed out on relevant multilingual safety works that are either not peer reviewed (yet) or published in other venues, such as ML conferences and workshops. 
Since it is a very fast moving field, the state of the field described in this paper represents a snapshot in time. We hope that if we ran an analysis like this in a year's time, the data would hopefully paint a more optimistic picture.

\textbf{Annotation accuracy} Inaccuracies in our annotations might have introduced imprecision in our measurements of the language gap. From our analysis of the inter-annotator agreement, we suspect that this would foremost affect the categorization of safety research topics, as the labels for these categories carry the most ambiguity. When papers do not state language coverage very prominently, such in the abstract or introduction or the experimental setup, it might lead to oversight in the annotation (reducing recall in annotations), depending how deeply an annotator reads the paper. However, we observe that especially those works that are investigating multilinguality in LLM safety as a primary angle, do state it explicitly, so we are confident we did not miss these.

\textbf{Research directions} We highlight three prominent future directions for multilingual safety research in our work, but we believe there are many other directions that are equally important for advancing safety and security of LLMs in global deployment. These include work on AI governance \citep{reuel2024open}, AI auditing \citep{birhane2024ai, ojewale2025towards},  hate speech detection \citep{nozza-2021-exposing}, multimodal AI safety \citep{dash2025aya,ji2025safe}, algorithmic designs \citep{zhao2025mpo}, reasoning models \citep{yong2025crosslingual,guan2024deliberative}, etc. Fundamentally, our work illuminates the substantial language disparity within current LLM safety research. Therefore, as researchers pursue diverse research directions on LLM safety, efforts on bridging this linguistic divide must remain central to ensuring equitable safeguards across the world's languages.

\section*{Acknowledgements and Disclosure}
We are grateful for feedback from Hellina Hailu Nigatu, M Saiful Bari, Cristina Menghini, Alham Fikri Aji, Pedro Ortiz Suarez, Victor Ojewale, Simran Khanuja, Arianna Muti,
Debora Nozza, Srishti Yadav, Jonas Kgomo, and Catherine Arnett on the early draft of our work. Disclosure: Stephen Bach is an advisor to Snorkel AI, a company that provides software
and services for data-centric artificial intelligence.

\bibliography{paper}

\end{document}